# A Study on the Effectiveness of Different Patch Size and Mouth Detection

Lim Huey Charn, Liyana Nuraini Rasid, Shahrel A. Suandi

Intelligent Biometric Group, School of Electrical and Electronic Engineering
Engineering Campus , Universiti Sains Malaysia
14300 Nibong Tebal, Pulau Pinang, MALAYSIA

*Abstract* – Template matching is one of the simplest methods used for eyes and mouth detection. However, it can be modified and extended to become a powerful tool. Since the patch plays a significant role in optimizing detection performance, a study on the influence of patch size and shape is carried out. The optimum patch size and shape is determined using the proposed method. Usually, template matching is also combined with other methods in order to improve detection accuracy. Thus, in this paper, the effectiveness of two image processing methods i.e. grayscale and Haar wavelet transform, when used with template matching are analyzed.

## I. INTRODUCTION

Since each human faces are unique, combinations of face features can help us distinguishing their identities and expressions. Thus, the purpose of this paper is to detect the location of the facial feature, specifically eyes and mouth, in a face region. Eyes and mouth are emphasized because these two features are the most notable recognition feature in a face, and they provide very important information that can be used for face recognition [1],[2]. Normally, by looking at humans' eyes alone, we can identify them correctly, given that we are familiar with them. Face feature detection is regarded as the crucial first step towards facial recognition and facial expression system. Hence, it can be utilized in different type of real-time application such as biometrics, video surveillance, human-machine interaction system, drowsy driver detection system [3], audio-visual speech processing [4], image retrieval (for automated process to image labeling) [5], visual reconstruction surgery [6] and etc.

There are lots of different approaches used by the researchers to detect eyes and mouth. However, in this paper, template matching method is applied to search for the location of the targeted object (i.e. eyes and mouth) in static images. Even though this method is simple, it can be modified to make it more robust and flexible. To modify it, two different features namely grayscale and Haar Wavelet are used with template matching. Thus, a system to evaluate the performance of both methods is developed. Also, since patch size is imperative in optimizing detection performance, suitable patch is determined by testing two different patch shapes (square and rectangular), and reducing the patch size gradually.

The paper is organized as follows: Section 2 briefly describes the existing template matching method, Section 3 gives full picture of the proposed method, and Section 4 explains the template matching process in details. Result of the study conducted is then shown in Section 5. In Section 6, the result are analyzed and discussed thoroughly. Finally, it ended up with conclusion in Section 7.

## II. EXISTING TEMPLATE MATCHING METHODS

Despite its simplicity, template matching suffers from a few limitations including high computational cost, sensitivity to noise, lighting condition, background noises and weak toward rotation [7],[8],[9]. Numerous researches had attempted to overcome this limitation by implementing different techniques to be combined with template matching.

Feris et al. [10] used a statistical skin-color model to detect face candidate regions and then, an efficient template matching scheme was applied to report the presence or absence of a face in these regions. Besides the template matching method, they also use hierarchical wavelet network to detect the facial feature. This method takes a coarse-to-fine approach to localize small features using cascading set of Gabor Wavelet Network features [11].

In Suandi et al. work, template matching method is applied to track human's eyes and mouth in real time. Good tracking result is obtained because accuracy of the system is increased with the additional of the energy minimization criterion and feature selective method [12].

In [13], face region from gray image is detected using an Ada Boosting algorithm. Then, the head tilt angle and the vertical location of two eyes is estimated using intensity valley and edge patterns. Next, candidate regions for irises are detected using a simple effective filter responding to the generic eye pattern. Finally, the eye is localized correctly using a fast deformable template matching algorithm.

Chai et al. [14] has proposed an algorithm to reduce the effect of moustache and beard in feature detection using template matching. Skin color information is used to locate the face region. Iris is then selected from valleys based on computed cost. For lips region detection, RGB color space is used, and then image processing method is applied to eliminate unwanted noise. Finally, template matching process is used to classify faces based on the extracted features.

In this paper, several alterations are made to overcome the template matching limitation. For example, to overcome high computational cost, multi-resolution template matching is done by creating an image pyramid (i.e. image hierarchies incorporating variable resolution). Small template is matched against small image and vice versa. The strongest matches will indicate us the location of our target object. Hierarchical template matching used in this study can save computation time and allows a considerable degree of insensitivity to noise.

## III. PROPOSED METHOD

Shown in Figure 1 is the flow chart of the proposed process. As mentioned before, image from two different processing techniques (i.e. grayscale and Haar Wavelet) are used for template matching. Figure 2 generally explains the template matching process, which is made up of the masking process, resizing image data, calculating and comparing correlation coefficient value.

### A. Face Model Partition

For the system developed, user is required to manually input four points indicating the face region (by mouse click). The selected image is being divided into three parts, i.e. an eye or a mouth in each partitioned area. From the geometrical constrain in face, we can assume that lower half of the face's height is the possible location for the mouth. On the other hand, the eye position can be assumed by dividing the upper face region by two, which the eye positions are in right and left sides respectively. Figure 3 illustrated the face partition model. This model is created to reduce the scanning time during the template matching process.

### B. Patch Size

Patch size plays a significant role in template matching. If the patch size is smaller than targeted object, the system will not be able to detect the right position for the object.

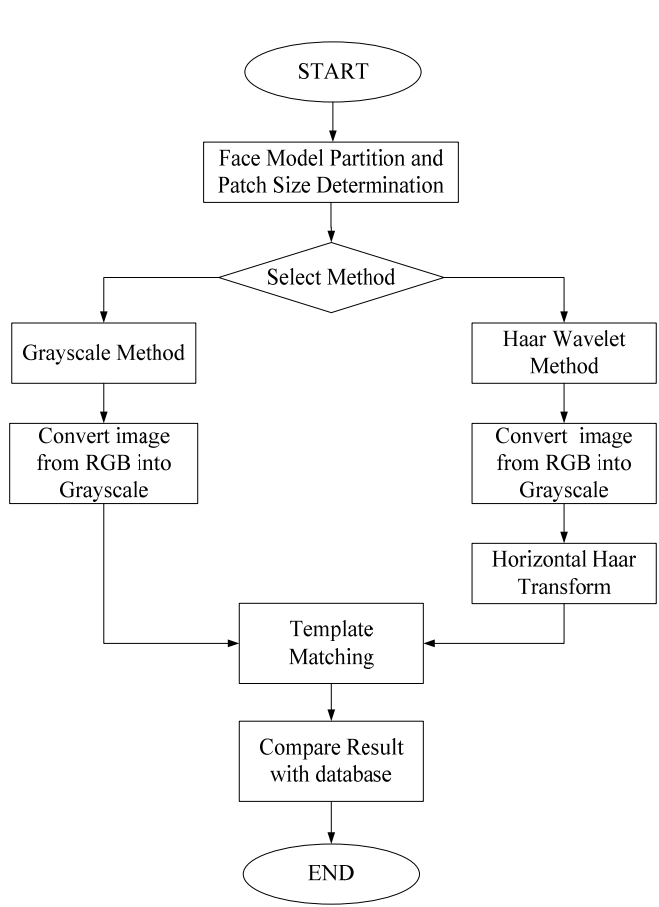

Figure 1. Flow chart of the eyes and mouth detection process

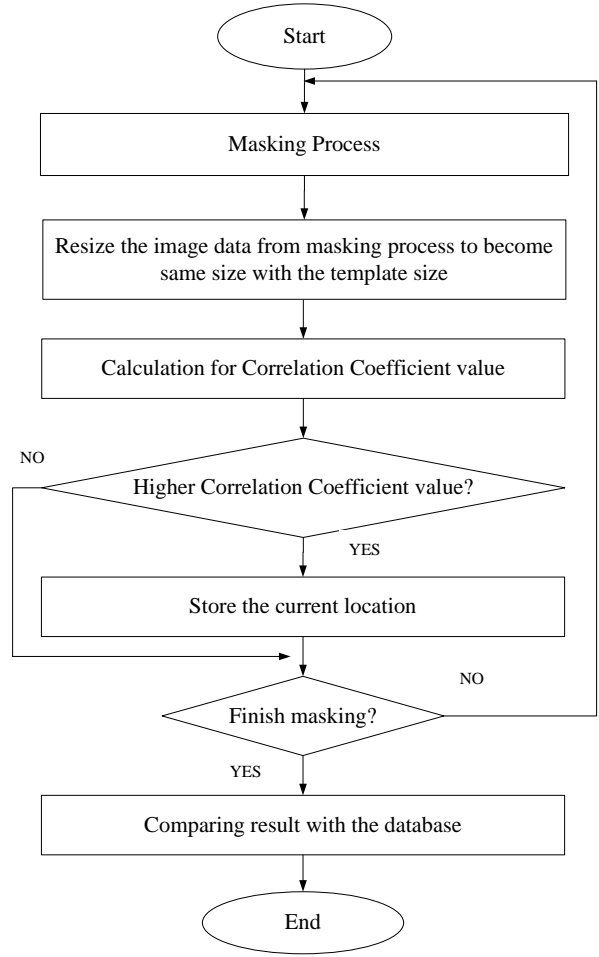

Figure 2. Flow chart for template matching process

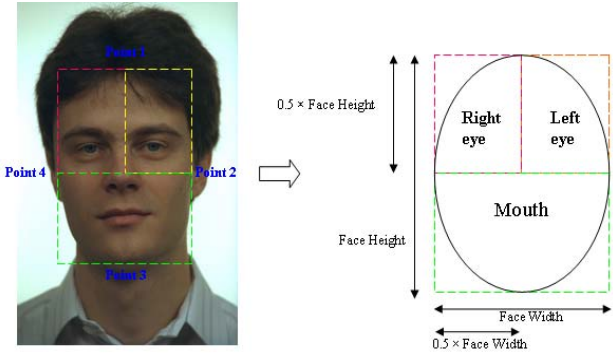

Figure 3.   Face model partition where the face region is divided into three sections i.e. right eye, left eye and mouth

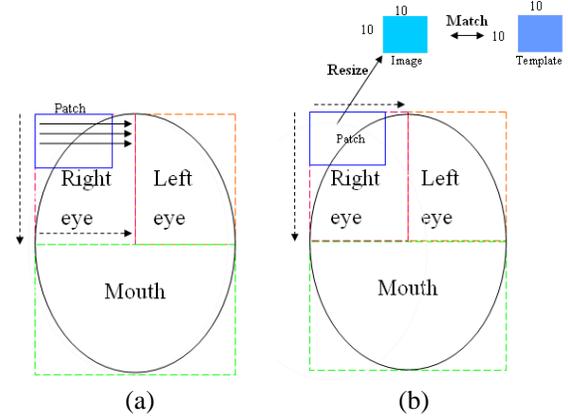

Figure 4.   (a) Patch scanning process and (b) Resize and match process

In this study, a square and a rectangular patch shape are used in order to test the detection accuracy when different patch size is utilized.

The square patch shape is selected because it facilitates the resize process. The square patch value is determined in Equation (1) and (2). Same square patch size is applied to all three features.

$$w_e = h_e = 0.5 \times w_f \qquad (1)$$

$$w_m = h_m = 0.5 \times w_f \qquad (2)$$

where

$w_e, h_e$     are eye patch width and height, respectively

$w_m, h_m$   are mouth patch width and height, respectively

$w_f$          is the face width

The value for the rectangular patch size is set by using equations below [12].

$$w_e = 0.5 \times w_f \qquad (3)$$

$$h_e = 0.8 \times w_e \qquad (4)$$

$$w_m = 0.6 \times w_f \qquad (5)$$

$$h_m = 0.8 \times w_m \qquad (6)$$

### C.   Image Conversion

As in Figure 1, for both methods, the color image is first converted into a grayscale image. Later, for the Haar method, Horizontal Haar Wavelet transform is applied onto the images.

There are three types of Haar wavelet transform i.e. Vertical Haar transform, Horizontal Haar Transform and Diagonal Haar Transform. Since eyes and mouth is located horizontally to the human face, Horizontal Haar Transform is applied to low pass the horizontal information and to high pass the vertical information. This method can be seen as a filter to reduce noise that influences the system performance. At the same time, this technique is able to make the horizontal edge clearer. Also, influence of hair and beard towards result can be reduced by using this technique.

### IV.   TEMPLATE MATCHING

In template matching, the template patch is scanned across the image and the highest correlation value is sought. Highest correlation value will indicate the location of the targeted object.

### A.   Eye Template

In this project, the template size dimension used is 10×10 pixels. To compute the template, a number of eye data is collected, resized into 10×10 pixels, and then the average pixel value of eye data is calculated pixel by pixel.

For example, twenty people's eye data with different dimension is taken. Then, the eye data is being resized into 10×10 pixels, resulting in twenty set of eye data in same dimension. The first pixel value of the template is determined by taking the average value of the first pixel from each eye data. The rest of the template pixel value is also obtained by the same manner.

Since there are two image processing methods in this project, different template is used for each method. For grayscale template, the image is converted from RGB to grayscale first, before going through the resizing and averaging process. The same manner is also applied to the Haar transformed image. For Haar template, the image is first converted into grayscale and then Horizontal Haar Transform is performed. Then, it will undergo the same resizing and averaging process as described before.

The effect of the template data is proportional to the number of eye data. The more the eye data taken for their averaging value, the better the template. However, there is a limitation number for the eye data. Too much eye data set will affect the effectiveness of the template. This issue will be addressed in our future paper.

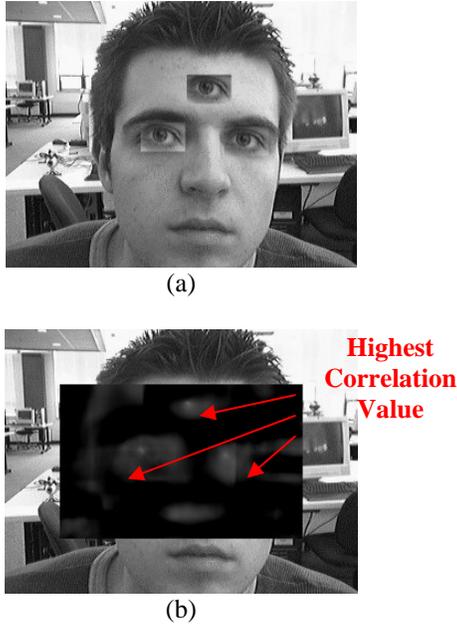

(a)

Highest
Correlation
Value

(b)

Figure 5. Detected result, (a) face with eye, (b) the highest correlation value representing the location for the center of the eye

TABLE I.    RECTANGULAR PATCH SIZES FOR EACH RESIZE PROCESS

|  | Left and Right Eye | Mouth |
|---|---|---|
| Original rectangular | $w_e = 0.5 \times w_f$<br>$h_e = 0.8 \times w_e$ | $w_m = 0.6 \times w_f$<br>$h_m = 0.8 \times w_m$ |
| After reduced 10% from original | $w_e^1 = 0.9 \times w_e$<br>$h_e^1 = 0.8 \times w_e^1$ | $w_m^1 = 0.9 \times w_m$<br>$h_m^1 = 0.8 \times w_m^1$ |
| After satreduced 20% from original | $w_e^2 = 0.8 \times w_e$<br>$h_e^2 = 0.8 \times w_e^2$ | $w_m^2 = 0.8 \times w_m$<br>$h_m^2 = 0.8 \times w_m^2$ |
| After reduced 30% from original | $w_e^3 = 0.7 \times w_e$<br>$h_e^3 = 0.8 \times w_e^3$ | $w_m^3 = 0.7 \times w_m$<br>$h_m^3 = 0.8 \times w_m^3$ |

TABLE II.    SQURE PATCH SIZES FOR EACH RESIZE PROCESS

|  | Left and Right Eye | Mouth |
|---|---|---|
| Original square | $w_e = h_e = 0.50 \times w_f$ | $w_m = h_m = 0.50 \times w_f$ |
| After 1st reduction | $w_e^1 = h_e^1 = 0.45 \times w_f$ | $w_m^1 = h_m^1 = 0.45 \times w_f$ |
| After 2nd reduction | $w_e^2 = h_e^2 = 0.45 \times w_f$ | $w_m^2 = h_m^2 = 0.55 \times w_f$ |

### B. Patch Scanning

The small blue colored frame in Figure 4a represents the patch. The patch is scanned through the feature region pixel by pixel from top left to bottom right. During the patch scanning process, the patch is moved around in the feature area to search for the highest correlation value. The output (i.e. correlation value) is highest at places where the image structure matches the template patch the best.

### C. Resize & Match

Information covered under the patch size is collected and the data is resized into the same size as the template i.e. 10x10 (refer to Figure 4b). After that, the correlation coefficient is calculated to determine the object location.

### D. Coefficient Correlation

Equation (7) from Gonzalez & Woods [15] is used to calculate correlation coefficient:

$$\gamma(x, y) = \frac{\sum_s \sum_t [f(s,t) - \overline{f}(s,t)][w(x+s, y+t) - \overline{w}]}{\left( \sum_s \sum_t [f(s,t) - \overline{f}(s,t)]^2 \sum_s \sum_t [w(x+s, y+t) - \overline{w}]^2 \right)^{1/2}}$$

where                                                              (7)

$\gamma(x, y)$    is the correlation coefficient

$f(x, y)$    is the pixel value at the coordinate (x, y)

$\overline{f}(x, y)$    is the average value of the region where the template and the image overlap at coordinate (x, y)

$w(x, y)$    is the template pixel value at coordinate (x, y)

$\overline{w}$    is the average value of the template

The template $w$ is moved across image $f$ in xy plane and the correlation for each point is calculated. The range for the correlation coefficient is between -1 and 1. It is independent of scale changes in the amplitude of $f$ and $w$.

The correlation coefficient value at the current location of the patch is compared with the previous position, as the patch move pixel by pixel within the feature region. If the correlation value for current patch location is higher than the previous location, the current coordinate of the patch is recorded. This coordinate will keep overwriting if the later position poses higher correlation coefficient value.

After finish scanning the related area, the highest correlation value representing the location of the target is obtained (refer Figure 5). Thus, position of the targeted feature is known and its will be displayed.

### E. Resize Patch Size

To obtain the most suitable patch size that provides the best detection result, the rectangular and square patch size is gradually reduced. This step is used to get the best ratio between patch size and the template size.

The rectangular patch sizes are reduced by 10%, 20% and 30% from the original sizes (refer to Table I) while for square patch, the sizes shown in Table II above are used.

## V. RESULTS

About 201 frontal images drawn from the FERET database are used for testing [16][17]. The 201 images consist of 119 faces with normal face, 41 faces with long front hair and 41 faces with spectacles. All the images are in the same size, which is 512 x 768 pixels. Result is verified based on the detected area of the eye or mouth (see judging criterion in Discussion). Since two methods are carried out in this project, the accuracy based on both methods is evaluated. Comparison of advantages and limitations of each method will be discussed in the next section.

Equation (8) is used to calculate the accuracy for each facial feature of the system.

$$\frac{\text{Accuracy for}}{\text{Each Feature}} = \frac{\text{Number of Successive Single Feature Detected}}{\text{Total Number of Image}} \times 100\%$$

(8)

$$\frac{\text{Average accuracy for}}{\text{Each Feature}} = \frac{NF + FLH + FS}{3}$$

(9)

where

NF      is accuracy of each feature for normal face

FLH     is accuracy of each feature for face with long front hair

FS      is accuracy of each feature for face with spectacles

Table III and IV show the result obtained from using rectangular and square patch, respectively, along with its accuracy graph in Figure 6 and 7.

### A. Image Type Analysis

Since there are three types of images used for testing, an analysis to judge the two proposed method on its ability to detect eyes and mouth on normal face image, face with long front hair and face with spectacles had been carried out separately.

From the result tested on 119 images with normal face in frontal pose, the accuracy for both methods is almost the same.

From the 201 images, 41 images consist of face with long front hair. Table VI below shows the result from the experiment. From the observation, Haar method performs better onto face image with long front hair. However, if the hair covered the eye brows, this will affect the result of the method. Figure 8 shows the detected result.

41 images consist of face with spectacles had been tested. Table VII shows the result from the experiment. Result in Figure 9 shows that the grayscale method performs better than the Haar method for face with spectacles.

## VI. DISCUSSION

### A. Judging Criterion

Judging criteria had been set to determine the accuracy of the system.   The result is considered as pass if the eye with eye

TABLE III.      RESULT FOR THE DIFFERENT RECTANGULAR PATCH

| Method | | Average Accuracy for each Face feature | | |
|---|---|---|---|---|
| | | *Left Eye* | *Right Eye* | *Mouth* |
| Accuracy of patch size with original size | Grayscale | 87.74% | 78.80% | 88.47% |
| | Haar | 90.46% | 95.37% | 90.10% |
| Accuracy of patch size after reduce 10% from original size | Grayscale | 71.37% | 68.69% | 70.31% |
| | Haar | 86.72% | 88.78% | 83.50% |
| Accuracy of patch size after reduce 20% from original size | Grayscale | 41.67% | 57.39% | 35.20% |
| | Haar | 62.83% | 54.45% | 41.87% |
| Accuracy of patch size after reduce 30% from original size | Grayscale | 21.37% | 15.94% | 0.81% |
| | Haar | 10.30% | 21.70% | 1.89% |

TABLE IV.      RESULT FOR THE DIFFERENT SQUARE PATCH

| Method | | Average Accuracy for each Face feature | | |
|---|---|---|---|---|
| | | *Left Eye* | *Right Eye* | *Mouth* |
| Eye Patch = 0.50 × Face width | Grayscale | 86.62% | 88.27% | 49.96% |
| Mouth Patch = 0.50 × Face width | Haar | 90.18% | 93.18% | 71.38% |
| Eye Patch = 0.45 × Face width | Grayscale | 63.85% | 73.02% | 24.32% |
| Mouth Patch = 0.45 × Face width | Haar | 82.00% | 76.38% | 40.39% |
| Eye Patch = 0.45 × Face width | Grayscale | 68.85% | 73.02% | 70.63% |
| Mouth Patch = 0.55 × Face width | Haar | 82.00% | 76.38% | 88.70% |

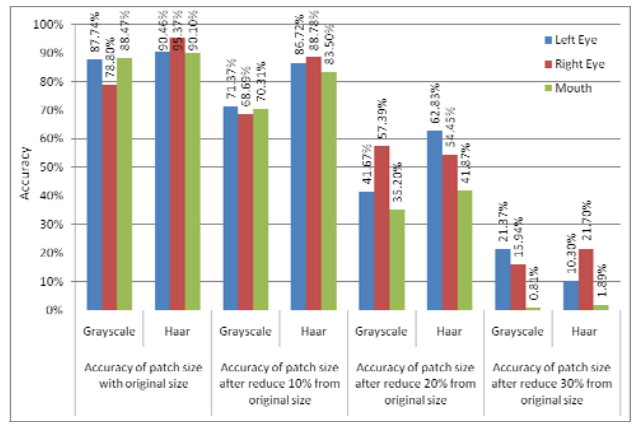

Figure 6.   Accuracy graph for the rectangular patch

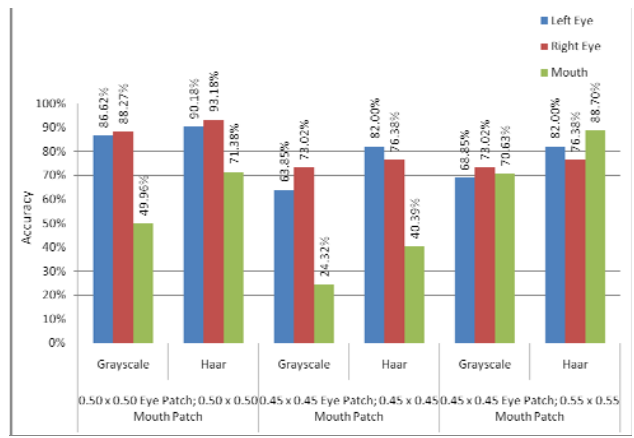

Figure 7.   Accuracy graph for the square patch

TABLE V.        RESULT FOR NORMAL FRONTAL POSE

| Method | Accuracy | | | Average Accuracy |
|---|---|---|---|---|
| | Left Eye | Right Eye | Mouth | |
| Grayscale | 94.95% | 94.95% | 87.39% | 92.43% |
| Haar | 95.79% | 98.31% | 87.39% | 93.83% |

TABLE VI.        RESULT FOR FACE WITH LONG FRONT HAIR

| Method | Accuracy | | Average Accuracy |
|---|---|---|---|
| | Left Eye | Right Eye | |
| Grayscale | 75.60% | 51.21% | 63.40% |
| Haar | 95.12% | 97.56% | 96.34% |

TABLE VII.        RESULT FOR FACE WITH SPECTACLE

| Method | Accuracy | | Average Accuracy |
|---|---|---|---|
| | Left Eye | Right Eye | |
| Grayscale | 92.68% | 90.24% | 91.46% |
| Haar | 80.48% | 90.24% | 85.36% |

TABLE VIII.        SUMMARY FOR THE BEST PATCH SIZE

| | Size |
|---|---|
| Left Eye | Rectangular original proposed size (using Haar method) |
| Right Eye | Rectangular original proposed size (using Haar method) |
| Mouth | Rectangular original proposed size (using Haar method) |

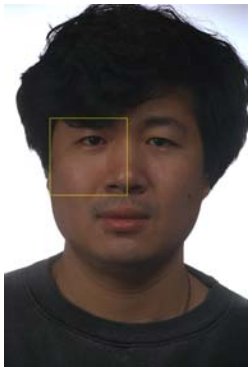

Figure 8.    Result for face with long front hair when eye brows is covered by hair

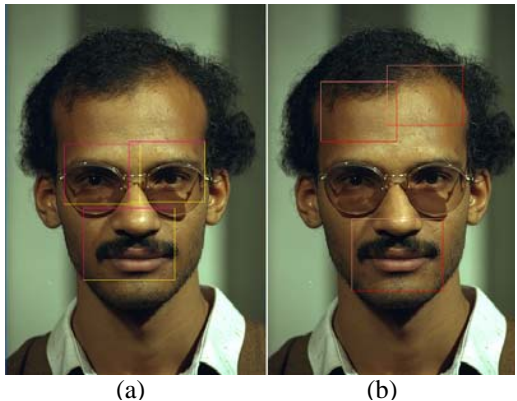

(a)                    (b)

Figure 9.    Result for face with spectacles using, (a) Grayscale Method, (b) Haar Method

brows or mouth with nostril is within the patch area. If the eye brows exceeds the patch area for a small number of pixels and it is not significant, the result is also considered as success. Similar judging criteria are applied to the mouth detection part.

Figure 10 and 11 illustrated the success and failure in judging eyes and mouth, respectively.

### B.  Patch Scanning Area

The patch scanning area also influences the detection result. There is possibility that the eye is not exactly located within the defined eye area, i.e. it might exceed few pixels from the defined area. Thus, larger scanning area for both eyes is needed (as shown in Figure 12). Besides, this larger scanning area can also improve the detection result for image with a hint of slanted pose.

From geometrical constraint, we can assume that even though mouth is located beneath the eye region, the mouth width is much smaller than the width from rightmost to leftmost eye. So, there is no possibility of mouth located outside that area. Thus, the width scanning area for mouth can be reduced in order to save scanning time.

### C.  Resizing Process

In template matching, if the data collected from the patch is smaller than the template size, it is hard for us to detect the location of our target accurately. However, if the data collected from the patch is much larger than the template size, the system is also not able to perform good detection result. Consequently, the resize process is needed to make both of the data compatible in the matching process.

### D.  The Effectiveness of the Method

Two methods are proposed to judge their accuracy on different types of face image.

Compared to grayscale method, Haar Transform method has a better detection result for face image with long front hair. It is because by using Horizontal Haar Transform, the horizontal intensity become more obvious compared to vertical and diagonal intensity.

Nevertheless, grayscale method is able to perform better in detecting feature in image of face with spectacles. This is because during resizing process, the noise created by the spectacle has been reduced. Thus, this method performs well for face with spectacles.

### E.  Comparison between the Two Proposed Patch Shape

Two different patch shapes (i.e. rectangular and square) has been used in this study. Figure 13 illustrated the highest accuracy for each feature based on the patch shape. As shown in the chart, the accuracy of the square patch is proven to be lower compared to the rectangular patch. The square patch is proposed due to the reason that it is more convenient to perform the resizing process. However, since eyes and mouth are rectangular, the template data taken is also originally rectangular. As a result, rectangular patch is able to perform better than square patch.

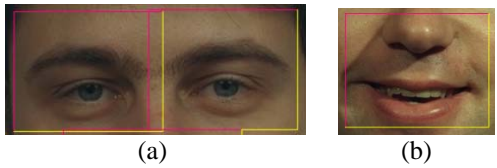

(a)                    (b)

Figure 10.  Successful result in (a) left and right eyes detection, and (b) mouth detection

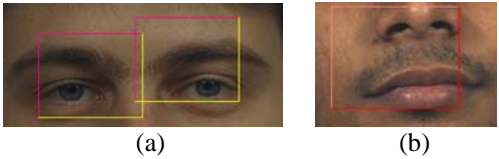

(a)                    (b)

Figure 11.  Failure result in (a) left and right eyes detection, and (b) mouth detection

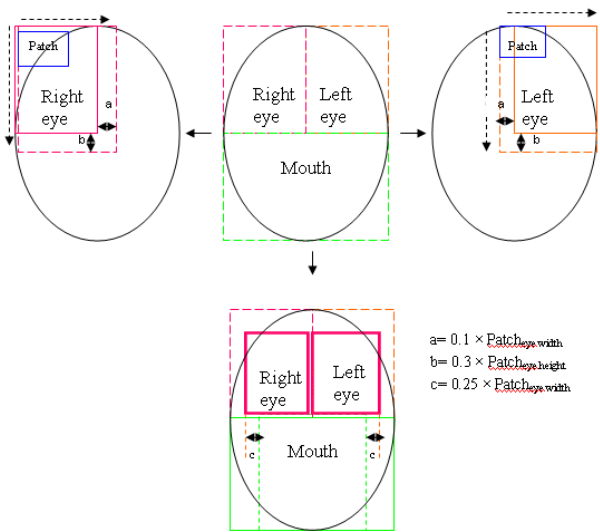

Figure 12.  Patch Scanning Area

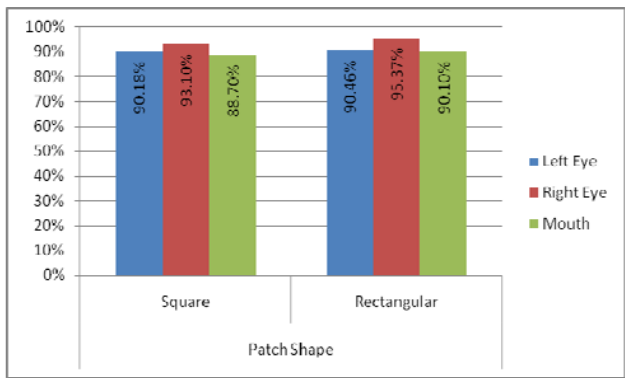

Figure 13.  Comparison of patch shape accuracy between a square and rectangular

### F.  The Patch Size

To find the most optimum patch size for template matching, the original patch size is reduced several times. From the observation, the accuracy of the system decreases when the patch size is being reduced from the original proposed size. Table VIII summarized the best result for each patch size with respect to the features.

### G.  Multiple Face Image

An experiment has been carried out to test the system response on multiple face images. The results are shown in Figure 14.

From the observation, both methods are able to perform eyes and mouth detection successfully (refer to Figure 14a and 14b). In Figure 14c, good detection result obtained for person A. However, left eye detection using grayscale method for person B and person C failed, because both face have long frontal hair. Black color frame in Figure 14d indicating the face model partition while the red frame indicating the detected facial feature. The detection result of person D failed due to the incorrect face model partitioning, which caused by the slanted face pose. Image processing technique like Ohtsu method might be required in order for the system to partition the slanted face model correctly.

## VII.  Conclusion

Since facial features detection has various applications, countless research had been carried out since around 20 years ago. Different strategies had been proposed to achieve robust detection. In this paper, eyes and mouth recognition system based on template matching method is developed.

A total of 201 images from the FERET database have been used to evaluate the accuracy of the proposed method. Since two different techniques i.e. grayscale and Horizontal Haar Transform are used, different templates are applied accordingly for template matching process. Three different classes of image (i.e. normal frontal pose, face with spectacles and face with long front hair) are used to analyze the accuracy of the methods.

Referring the result obtained from the study, grayscale method is proven to perform better for face with spectacles. On the other hand, Horizontal Haar method performs more efficiently for face with long front hair rather than grayscale method. Both of the methods have their own advantage and limitation due to different face type in an image. Thus, the accuracy of the system can be improved by applying both methods onto the same image. The location with better correlation value will be considered as the best location for the targeted object.

Since both patch and template play a significant role in template matching method, the best patch size and shape is determined by applying both rectangular and square patch, and reducing its size gradually. Based from the results, rectangular patch with the original proposed size perform better than the rest. This is because the template itself is constructed using rectangular shape eyes and mouth data, and thus, rectangular

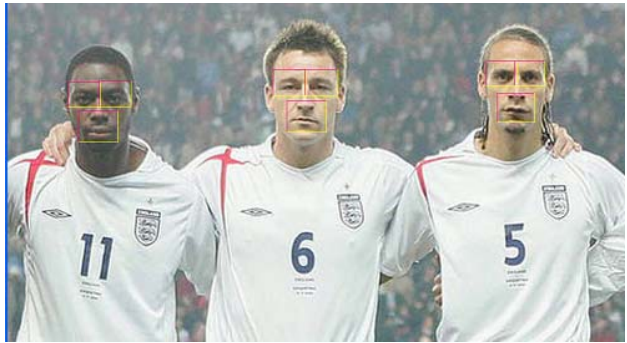

(a)

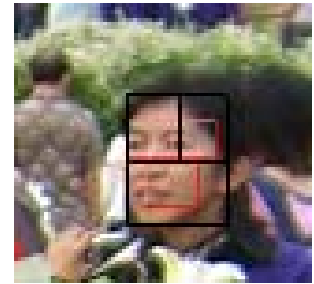

(d)

Figure 14. Result of multiple face image detectiion (a) Using grayscale method, (b) Using Haar method, (c) Failure in eyes detection due to long front hair and (d) Detection on person D

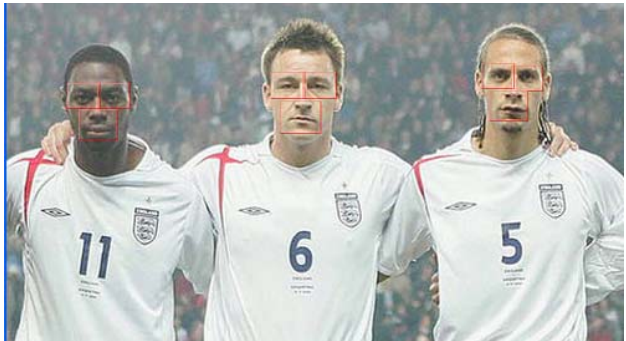

(b)

patch matched better. Also, from the study, the detection accuracy decrease with patch size. The optimum patch size in this case is the original size.

From the effect of the method and patch size, we can conclude that this view-based technique is very sensitive and it depends on the input to get our desired output. Further work can be carried out to increase its immunity. Cascade method, energy minimization technique and feature selective method can be added to the system in order to increase the system robustness.

ACKNOWLEDGMENT

This work is supported by both USM Fellowship and Universiti Sains Malaysia Short Term Grant under project code no. 304/PELECT/6035270.

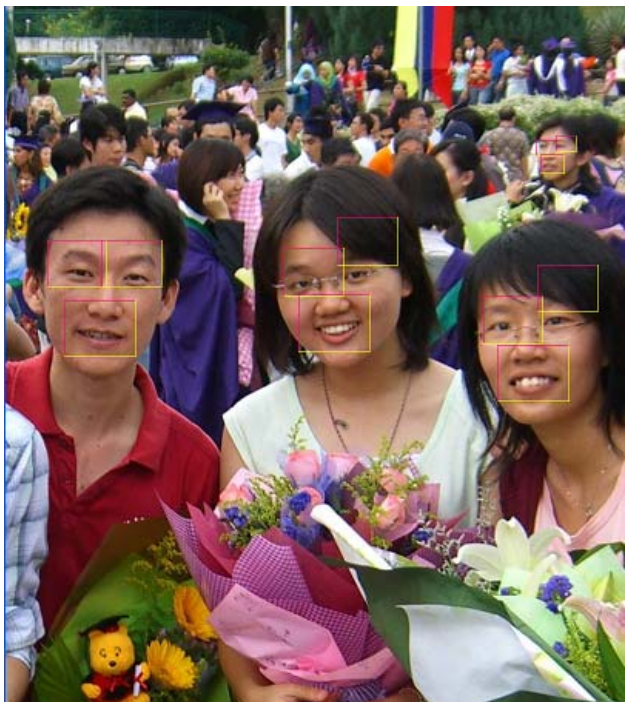

(c)

REFERENCES

[1] N. D. Haig, "Exploring recognition with interchanged facial features", Perception, vol. 15, no. 3, pp. 235-247, February 1986.

[2] P. Sinha, B. Balas, Y. Ostrovsky and R. Russell, "Face recognition by humans: nineteen results all computer vision researchers should know about", Proceedings of the IEEE Vol. 94, pp. 1948–1962, January 2007.

[3] N. Parmar, "Drowsy Driver Detection System", Engineering Design Project Thesis, Ryerson University, 2002.

[4] S. Lucey, S. Sridharan, and V. Chandran, "Improved Facial-Feature Detection for AVSP via Unsupervised Clustering and Discriminant Analysis" EURASIP Journal on Applied Signal Processing, 2003, vol. 3, pp. 264-275, 2003.

[5] J. D. Brand, "Visual speech for speaker recognition and robust face detection", PhD Thesis, University of Wales Swansea, 2001.

[6] S. A. Rabi and P. Aarabi, "Face Fusion: An Automatic Method For Virtual Plastic Surgery", 9th International Conference on Information Fusion, pp. 1-7, July 2006.

[7] R. Brunelli, Template Matching Techniques in Computer Vision; Theory and Practice, Wiley & Sons, 2009.

[8] S. A. Suandi, S. Enokida and T. Ejima, "EMoTracker: Eyes and mouth tracker based on energy minimization", 4th Indian Conference on Computer Vision, Graphics and Image Processing 2004 (ICVGIP 2004), pp. 269-274, 2004.

[9] R. Brunelli and T. Poggio. "Face Recognition: Features versus Templates", IEEE Transactions on Pattern Analysis and Machine Intelligence, vol. 15, no.10, pp. 1042-1052, 1993

[10] R. S. Feris, T. E. Campos, and R. M. C. Junior, "Detection and tracking of facial features in video sequences", Lecture Notes in Artificial Intelligence, vol. 1793, pp. 197-206, April 2000.

[11] R. S. Feris, J. Gemmell, J., K. Toyama, "Facial Feature Detection Using A Hierarchical Wavelet Face Database", Technical report in Microsoft Research, 2002.

[12] S. A. Suandi, S. Enokida and T. Ejima, "An Extended Template Matching Technique for Tracking Eyes and Mouth in Real-Time", 3rd IASTED International Conference on Visual, Imaging and Image Processing (VIIP 2003), pp. 815-820, 2003.

[13] Choi et al., "Reliable and fast eye detection", Proceedings of the 9th WSEAS International Conference on Signal Processing, Computational Geometry and Artificial Vision, pp. 78-82, 2009.

[14] T. Y. Chai, M.Rizon, S. S. Woo and C. S. Tan, "Facial Features for Template Matching Based Face Recognition", American Journal of Engineering and Applied Sciences, vol. 3, no. 1, 899-903, 2010.

[15] R. C. Gonzalez, and R. E. Woods, Digital Image Processing, 2nd edn, Prentice Hall Inc., New Jersey, 2002

[16] P. J. Phillips, H. Moon, S. A. Rizvi, and P. J. Rauss, "The FERET evaluation methodology for face recognition algorithms", IEEE Trans. Pattern Analysis and Machine Intelligence, vol.22, pp. 1090-1104, 2004.

[17] P. I. Phillips, H. Wechsler, J. Huang, and P. J. Rauss, "The FERET database and evaluation procedure for face recognition algorithm", Image and vision computing J, vol.16,no.5, pp. 295-306, 1998.